\pdfoutput=1

\documentclass[11pt]{article}

\usepackage{acl}
\usepackage{listings}
\usepackage{url}
\usepackage{booktabs, multirow}
\usepackage{arydshln}
\usepackage{enumitem}

\usepackage{times}
\usepackage{latexsym}
\usepackage{graphicx}
\usepackage[T1]{fontenc}

\definecolor{codegreen}{rgb}{0,0.6,0}
\definecolor{codegray}{rgb}{0.5,0.5,0.5}
\definecolor{codepurple}{rgb}{0.58,0,0.82}
\definecolor{backcolour}{rgb}{0.95,0.95,0.92}

\lstdefinestyle{mystyle}{
    backgroundcolor=\color{backcolour},   
    commentstyle=\color{codegreen},
    keywordstyle=\color{magenta},
    numberstyle=\tiny\color{codegray},
    stringstyle=\color{codepurple},
    basicstyle=\ttfamily\footnotesize,
    breakatwhitespace=false,         
    breaklines=true,                 
    captionpos=b,                    
    keepspaces=true,                 
    numbers=none,                    
    numbersep=5pt,                  
    showspaces=false,                
    showstringspaces=false,
    showtabs=false,                  
    tabsize=2,
}

\usepackage[utf8]{inputenc}

\usepackage{microtype}

\usepackage{inconsolata}

\title{LitLLM: A Toolkit for Scientific Literature Review}

\author{Shubham Agarwal*\textsuperscript{1,2,3}, Gaurav Sahu*\textsuperscript{1}, Abhay Puri*\textsuperscript{1}, Issam H. Laradji\textsuperscript{1,4}, \\
\textbf{Krishnamurthy DJ Dvijotham\textsuperscript{1}, Jason Stanley\textsuperscript{1}, Laurent Charlin\textsuperscript{2,3,5}, Christopher Pal\textsuperscript{1,2,5}} \\[1ex]
\textsuperscript{1}ServiceNow Research, \textsuperscript{2}Mila - Quebec AI Institute, \textsuperscript{3}HEC Montreal, Canada\\
\textsuperscript{4}UBC, Vancouver, Canada, \textsuperscript{5}Canada CIFAR AI Chair, \textsuperscript{6}University of Waterloo\\[1ex]
\texttt{Correspondence: \{shubham.agarwal, gaurav.sahu\}@mila.quebec
} \\
Journal version: \url{https://openreview.net/forum?id=heeJqQXKg7} \\
\thanks{Equal contribution.}
}

\begin{document}
\maketitle
\begin{abstract}
Conducting literature reviews for scientific papers is essential for understanding research, its limitations, and building on existing work. It is a tedious task which makes an automatic literature review generator appealing. Unfortunately, many existing works that generate such reviews using Large Language Models (LLMs) have significant limitations. They tend to hallucinate---generate non-factual information---and ignore the latest research they have not been trained on. To address these limitations, we propose a toolkit that operates on Retrieval Augmented Generation (RAG) principles, specialized prompting and instructing techniques with the help of LLMs.
Our system first initiates a web search to retrieve relevant papers by summarizing user-provided abstracts into keywords using an off-the-shelf LLM. Authors can enhance the search by supplementing it with relevant papers or keywords, contributing to a tailored retrieval process. Second, the system re-ranks the retrieved papers based on the user-provided abstract. Finally, the related work section is generated based on the re-ranked results and the abstract. There is a substantial reduction in time and effort for literature review compared to traditional methods, establishing our toolkit as an efficient alternative.
Our project page including the demo and toolkit can be accessed here: \url{https://litllm.github.io}.
\end{abstract} 


\section{Introduction}

Scientists have long used NLP systems like search engines to find and retrieve relevant papers. Scholarly engines, including Google Scholar, Microsoft Academic Graph, and Semantic Scholar, provide additional tools and structure to help researchers further. Following recent advances in large language models (LLMs), a new set of systems provides even more advanced features. For example, Explainpaper\footnote{\url{https://www.explainpaper.com/}} helps explain the contents of papers, and Writefull\footnote{\url{https://x.writefull.com/}} helps with several writing tasks, including abstract and title generation. There are, of course, many other tasks where similar technologies could be helpful. 

Systems that help researchers with literature reviews hold promising prospects. The literature review is a difficult task that can be decomposed into several sub-tasks, including retrieving relevant papers and generating a related works section that contextualizes the proposed work compared to the existing literature. 
It is also a task where factual correctness is essential. In that sense, it is a challenging task for current LLMs, which are known to hallucinate. Overall, creating tools to help researchers more rapidly identify, summarize and contextualize relevant prior work could significantly help the research community.

Recent works explore the task of literature review in parts or in full. For example, \citet{lu-etal-2020-multi-xscience} proposes generating the related works section of a paper using its abstract and a list of (relevant) references.
Researchers also look at the whole task and build systems using LLMs like ChatGPT for literature review~\cite{haman2023using, huang2023role}. While these LLMs tend to generate high-quality text, they are prone to hallucinations~\cite{Athaluri2023ExploringTB}. For example, the Galactica system was developed to reason about scientific knowledge \citep{taylor2022galactica}. While it outperforms contemporary models on various scientific tasks, it generates made-up content like inaccurate citations and imaginary papers.\footnote{see e.g., \href{https://venturebeat.com/ai/what-meta-learned-from-galactica-the-doomed-model-launched-two-weeks-before-chatgpt/}{What Meta Learned from Galactica}} 


\begin{figure*}
\centering
\includegraphics[width=\linewidth]{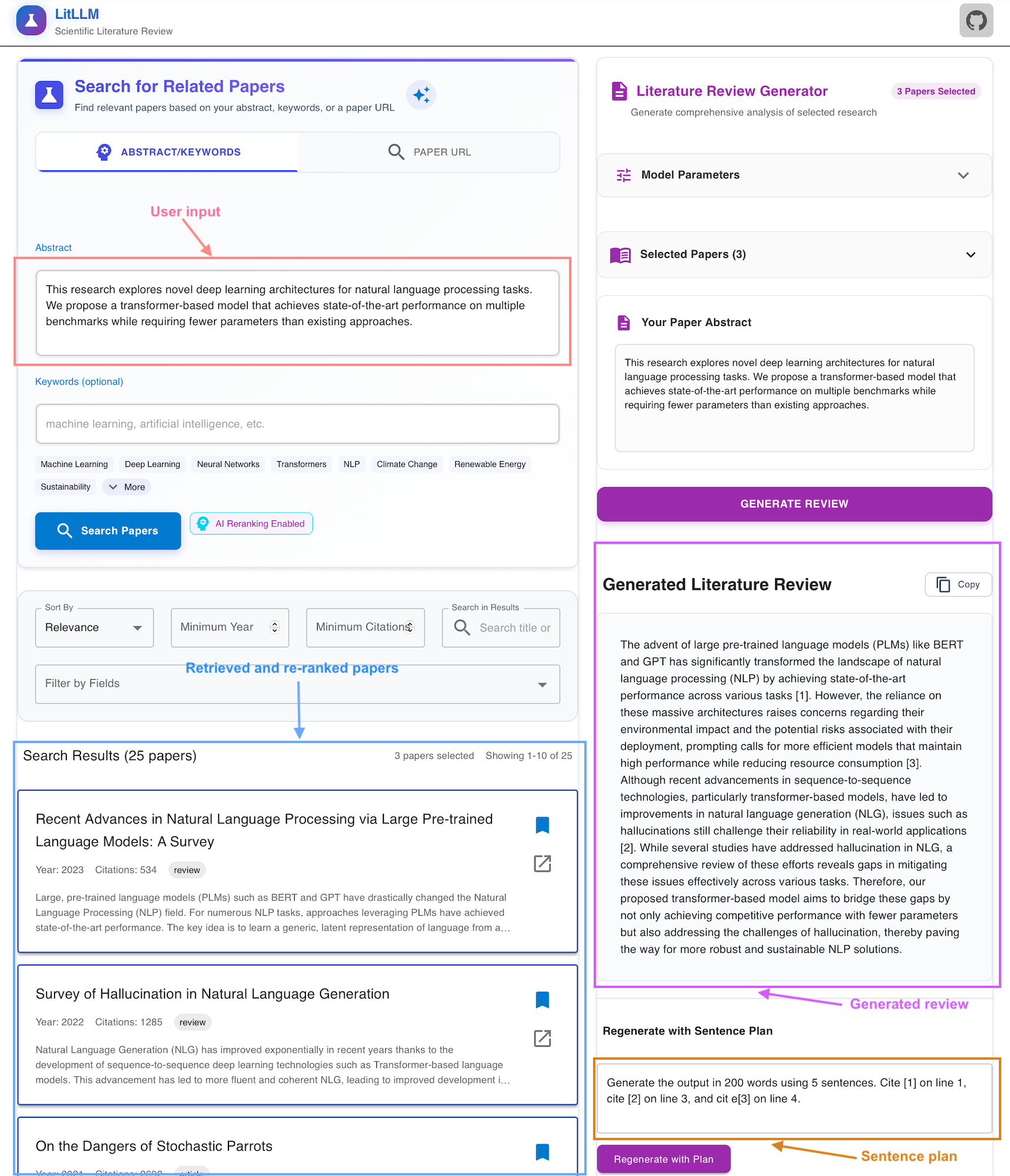}
\caption{LitLLM Interface. Our system works on the Retrieval Augmented Generation (RAG) principle to generate the literature review grounded in retrieved relevant papers. User needs to provide the abstract in the textbox (in purple) and press send to get the generated related work (in red). First, the abstract is summarized into keywords (Section \ref{sec:retrieval}), which are used to query a search engine. Retrieved results are re-ranked (in blue) using the Paper Re-Ranking module (Section \ref{sec:rerank}), which is then used as context to generate the related work (Section \ref{sec:generation}). Users could also provide a sentence plan (in green) according to their preference to generate a concise, readily usable literature review (See Section \ref{subsec:plan}).}
\label{fig:pull-figure}
\end{figure*}

As a step forward, we explore retrieval-augmented-generation (RAG) to improve factual correctness~\cite{lewis2020retrieval}. The idea is to use the retrieval mechanism to obtain a relevant list of existing papers to be cited which provides relevant contextual knowledge for LLM based generation.



LitLLM is an interactive tool to help scientists write the literature review or related work section of a scientific paper starting from a user-provided abstract (see Figure~\ref{fig:pull-figure}).
The specific objectives of this work are to create a system to help users navigate through research papers and write a literature review for a given paper or project. Our main contributions are:
\begin{itemize}[noitemsep,topsep=0pt]
    \item We provide a system based on a modular pipeline that conducts a literature review based on a user-proposed abstract.
    \item We use Retrieval Augmented Generation (RAG) techniques to condition the generated related work on factual content and avoid hallucinations using multiple search techniques. 
    \item We incorporate sentence-based planning to promote controllable generation. 
\end{itemize}

\begin{figure}
\centering
\includegraphics[width=\linewidth]{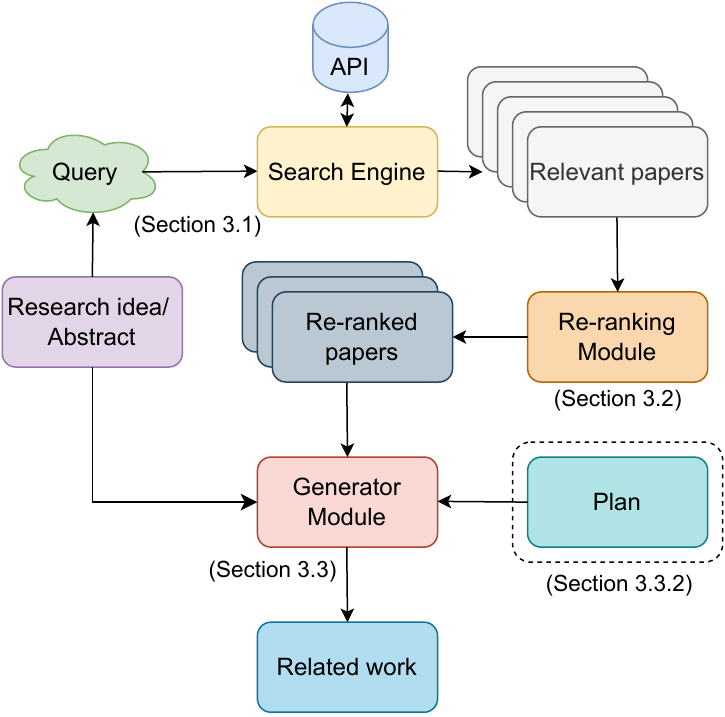}
\caption{Schematic diagram of the modular pipeline used in our system. In the default setup, we summarize the research abstract into a keyword query, which is used to retrieve relevant papers from an academic search engine. We use an LLM-based reranker to select the most relevant paper relative to the provided abstract. Based on the re-ranked results and the user-provided summary of their work, we use an LLM-based generative model to generate the literature review, optionally controlled by a sentence plan.}
\label{fig:flow}
\end{figure} 

\section{Related Work}

LLMs have demonstrated significant capabilities in storing factual knowledge and achieving state-of-the-art results when fine-tuned on downstream Natural Language Processing (NLP) tasks ~\cite{lewis2020retrieval}. 

However, they also face challenges such as hallucination, outdated knowledge, and non-transparent, untraceable reasoning processes ~\cite{huang2023survey,gao2023retrieval, li2024dawn}. These limitations have motivated the development of RAG (Retrieval Augmented Generation), which incorporates knowledge from external databases to enhance the accuracy and credibility of the models, particularly for knowledge-intensive tasks ~\cite{gao2023retrieval}. RAG has emerged as a promising solution to the challenges faced by LLMs. It synergistically merges LLMs' intrinsic knowledge with the vast, dynamic repositories of external databases ~\cite{gao2023retrieval}. This approach allows for continuous knowledge updates and integration of domain-specific information in an attempt to limit the effect of outdated knowledge. The proposed work builds upon the advancements around RAG to provide a more efficient solution for academic writing.

On the other hand, there has been a notable emphasis on utilizing Large Language Models (LLMs) for tasks related to information retrieval and ranking ~\cite{zhu2023large}. The work by~\citet{sun2023chatgpt} leverages generative LLMs such as ChatGPT and GPT-4 for relevance ranking in information retrieval, demonstrating that these models can deliver competitive results to state-of-the-art supervised methods. \citet{pradeep2023rankzephyr,pradeep2023rankvicuna} introduce different open-source LLM for listwise zero-shot reranking, further motivating the proposed approach of using LLMs for reranking in our work.

The exploration of large language models (LLMs) and their zero-shot abilities has been a significant focus in recent research. For instance, one study investigated using LLMs in recommender systems, demonstrating their promising zero-shot ranking abilities, although they struggled with the order of historical interactions and position bias ~\cite{Hou2023LargeLM}. Another study improved the zero-shot learning abilities of LLMs through instruction tuning, which led to substantial improvements in performance on unseen tasks ~\cite{Wei2021FinetunedLM}. A similar approach was taken to enhance the zero-shot reasoning abilities of LLMs, with the introduction of an autonomous agent to instruct the reasoning process, resulting in significant performance boosts ~\cite{Crispino2023AgentIL}. The application of LLMs has also been explored in the context of natural language generation (NLG) assessment, with comparative assessment found to be superior to prompt scoring ~\cite{Liusie2023LLMCA}. In the domain of Open-Domain Question Answering (ODQA), a Self-Prompting framework was proposed to utilize the massive knowledge stored in LLMs, leading to significant improvements over previous methods ~\cite{Li2022SelfPromptingLL}. Prompt engineering has been identified as a key technique for enhancing the abilities of LLMs, with various strategies being explored ~\cite{Shi2023PromptSO}.\footnote{This paragraph was generated using our platform with some minor modifications based on a slightly different version of our abstract.}

\section{Pipeline}
Figure \ref{fig:flow} provides an overview of the pipeline. The user provides a draft of the abstract or a research idea. We use LLM to first summarize the abstract in keywords that can be used as a query for search engines. Optionally, the users could provide relevant keywords to improve search results. This query is passed to the search engine, which retrieves relevant papers with the corresponding information, such as abstracts and open-access PDF URLs. These retrieved abstracts with the original query abstract are used as input to the other LLM Re-ranker, which provides a listwise ranking of the papers based on the relevance to the query abstract. These re-ranked abstracts with the original query are finally passed to the LLM generator, which generates the related work section of the paper. Recently, \citet{agarwal2024llms} showed that prompting the LLMs with the sentence plans results in reduced hallucinations in the generation outputs. These plans contain information about the number of sentences and the citation description on each line, providing control to meet author preferences. We include this sentence-based planning in the LLM generator as part of this system. In the following, we provide more details about each of the modules.
\begin{figure}[ht]
\centering
\includegraphics[width=0.9\linewidth]{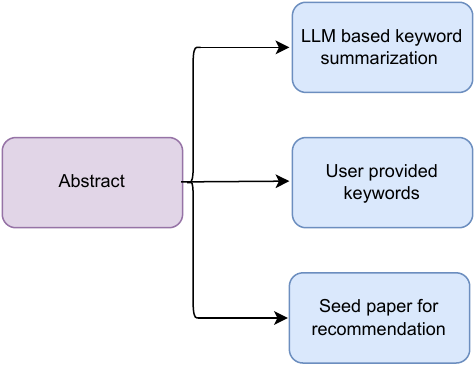}
\caption{Different retrieval strategies as discussed in Section \ref{sec:retrieval}}
\label{fig:strategies}
\end{figure}

\subsection{Paper Retrieval Module}
\label{sec:retrieval}
\begin{figure}[ht]
\centering
\includegraphics[width=\linewidth]{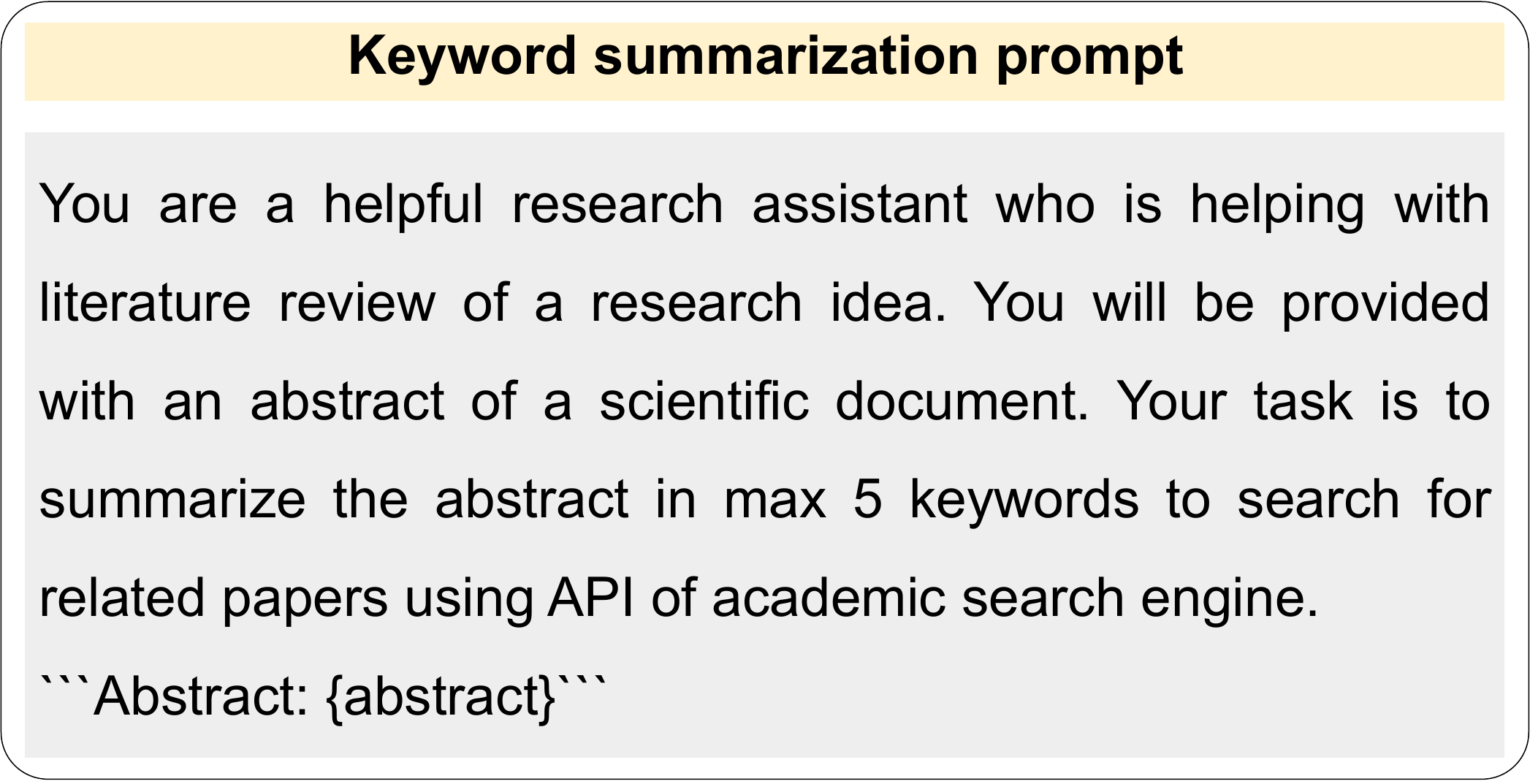}
\caption{Prompt used to summarize the research idea by LLM to search an academic engine}
\label{fig:summarization-prompt}
\end{figure}

In our toolkit, we retrieve relevant papers using the Semantic Scholar API~\cite{kinney2023semantic} and OpenAlex API~\cite{priem2022openalex}. 
Other platforms could be used, but the S2 and OpenAlex platforms are well-adapted to this use case. Combined, they constitue a large-scale academic corpus comprising 300M+ metadata records across multiple research areas, providing information about papers' metadata, authors, paper embedding, etc. The S2 Recommendations API also provides relevant papers similar to any seed paper. Figure \ref{fig:strategies} shows our system's different strategies. We describe these three settings that we use to search for references:

\begin{itemize}[noitemsep,topsep=0pt]
    \item User provides an abstract or a research idea (roughly the length of the abstract). We prompt an LLM (see Figure \ref{fig:summarization-prompt}) to summarize this abstract in keywords which can be used as a search query with most APIs.
    \item Users can optionally also provide keywords that can improve search results. This is similar (in spirit) to how researchers search for related work with a search engine. This is particularly useful in interdisciplinary research, and authors would like to include the latest research from a particular domain, which could not be captured much in the abstract.
    \item Lastly, any seed paper the user finds relevant enough to their idea could be used with the Recommendations API from search engines to provide other closely related papers. 
\end{itemize}

\subsection{Paper Re-Ranking Module}
\label{sec:rerank}

\begin{figure}[ht]
\centering
\includegraphics[width=\linewidth]{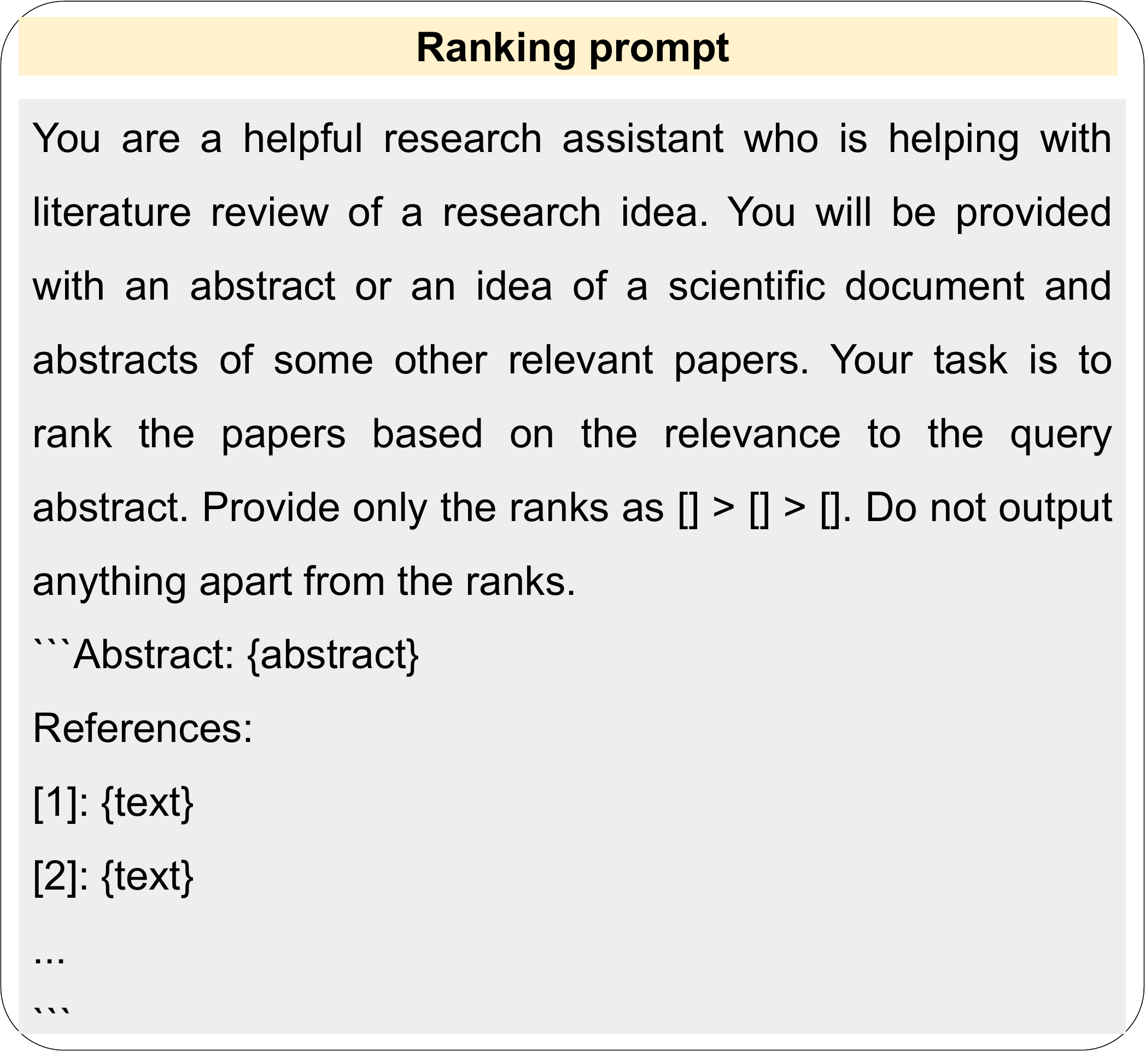}
\caption{Ranking prompt based on the permutation generation method}
\label{fig:ranking-prompt}
\end{figure}

Recent efforts have explored the application of proprietary LLMs for ranking ~\cite{sun2023chatgpt,ma2023zero} as well as open-source models like ~\cite{pradeep2023rankvicuna,pradeep2023rankzephyr}. These approaches provide a combined list of passages directly as input to the model and retrieve the re-ordered ranking list ~\cite{zhang2023rank}. Typically, a retriever first filters top-k potential candidates, which are then re-ranked by an LLM to provide the final output list. In our work, we explore the instructional \emph{permutation generation} approach~\cite{sun2023chatgpt} where the model is prompted to generate a permutation of the different papers in descending order based on the relevance to the user-provided abstract, and \textit{debate-ranking with attribution}~\cite{rahaman2024language,agarwal2024llms} where an LLM is prompted to (1) generate arguments for and against including the candidate paper and (2) output a final probability of including the candidate based on the arguments. Figure \ref{fig:ranking-prompt} showcases the prompt we used for LLM-based re-ranking. 

\begin{figure}[ht]
\centering
\includegraphics[width=\linewidth]{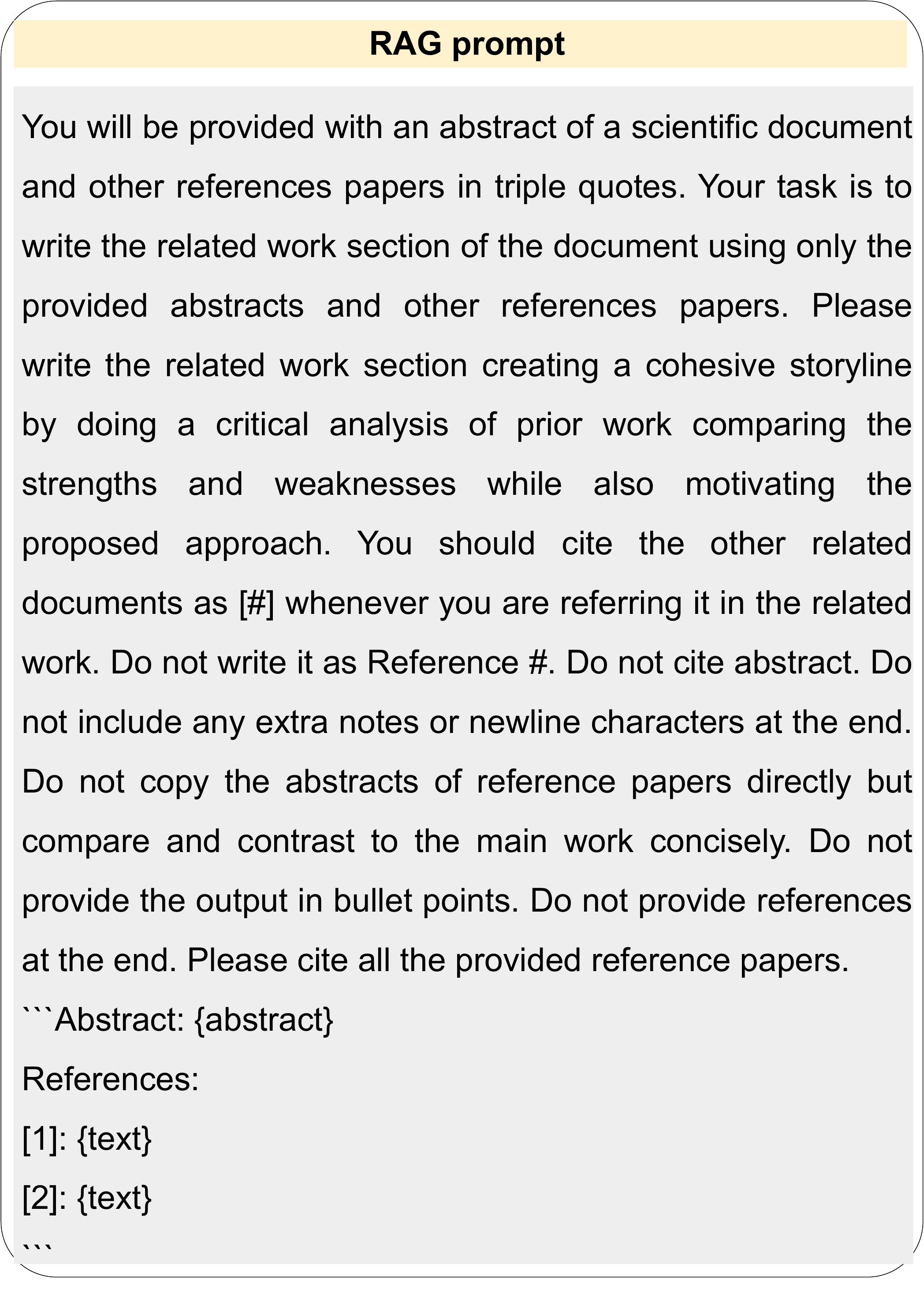}
\caption{Prompt for Retrieval Augmented Generation}
\label{fig:generation-prompt}
\end{figure} 

\subsection{Summary Generation Module}
\label{sec:generation}
We explore two strategies for generation: (1) Zero-shot generation and (2) Plan-based generation, which relies on sentence plans for controllable generation, described in the following 
\subsubsection{Zero-shot generation}
\label{subsec:zero}

While LLMs can potentially search and generate relevant papers from their parametric memory and trained data, they are prone to hallucinating and generating non-factual content. Retrieval augmented generation, first introduced in ~\citet{parvez2021retrieval} for knowledge tasks, addresses this by augmenting the generation model with an information retrieval module. 
The RAG principles have been subsequently used for dialogue generation in task-oriented settings~\cite{thulke2021efficient}, code generation~\cite{liu2020retrieval,parvez2021retrieval} and product review generation~\cite{kim2020retrieval}.
RAG drastically reduces hallucinations in the generated output~\cite{gao2023retrieval,tonmoy2024comprehensive}. 

Our work builds upon the principles of RAG, where we retrieve the relevant papers based on the query and augment them as context for generating the literature review. This also allows the system to be grounded in the retrieved information and be updated with the latest research where the training data limits the parametric knowledge of the LLM. Figure \ref{fig:generation-prompt} shows our system's prompt for effective Retrieval Augmented Generation (RAG).

\subsubsection{Plan based generation}
\label{subsec:plan}
To get the best results from LLM, recent research shifts focus on designing better prompts (Prompt Engineering) including 0-shot chain-of-thought prompting ~\cite{kojima2022large,zhou2022large}, few-shot prompting ~\cite{brown2020gpt3} techniques, few-shot Chain-of-thought prompting ~\cite{weichain} and in-context prompting ~\cite{li2021prefix,qin2021learning}. However, the longer context of our problem statement (query paper and multiple relevant papers) hinders the application of these techniques for response generation. 

We utilized sentence plan-based prompting techniques drawing upon insights from the literature of traditional modular Natural Language Generation (NLG) pipelines with intermediary steps of sentence planning and surface realization~\cite{reiter1997building,stent-etal-2004-trainable}. These plans provide a sentence structure of the expected output, which efficiently guides the LLM in generating the literature review in a controllable fashion as demonstrated in concurrent work \citep{agarwal2024llms}. Figure~\ref{fig:plan-generation-prompt} (in Appendix) shows the prompt for plan-based generation with an example template as:

\vspace{2mm}
\noindent\fbox{%
    \parbox{0.99\linewidth}{%
\footnotesize{\texttt{{Please generate \{num\_sentences\} sentences in \{num\_words\} words. Cite \{cite\_x\} at line \{line\_x\}. Cite \{cite\_y\} at line \{line\_y\}.
        }}}
    }%
}\\

\section{Implementation Details}

We build our system using React, which provides a nice interface to quickly and efficiently build system demos.
We query the Semantic Scholar API available through the Semantic Scholar Open Data Platform \citep{lo-etal-2020-s2orc,kinney2023semantic} to search for the relevant papers. Specifically, we use the Academic Graph\footnote{\url{https://api.semanticscholar.org/api-docs/graph}} and Recommendations\footnote{\url{https://api.semanticscholar.org/api-docs/recommendations}} API endpoint.
For OpenAlex, we query the search endpoint.~\footnote{\url{https://api.openalex.org/works}}
In this work, we use OpenAI API\footnote{\url{https://platform.openai.com/docs/guides/gpt}} to generate results for LLM using GPT-3.5-turbo and GPT-4 model. At the same time, our modular pipeline allows using any LLM (proprietary or open-sourced) for different components. We also allow the end-user to sort the retrieved papers by relevance (default S2 results), citation count, or year. 

\section{User Experience}
As a preliminary study, we provided access to our user interface to 5 different researchers who worked through the demo to write literature reviews and validate the system's efficacy. We also provide an example in the demo with an abstract for a quick start. Particularly, the users found the 0-shot generation to be more informative about the literature in general while the plan-based generation to be more accessible and tailored for their research paper, as also evident in our demo video.\footnote{\url{https://youtu.be/E2ggOZBAFw0}}. Table \ref{table:qualitative} (in Appendix) shows the output-related work for a recent paper~\cite{li2023dall} that was randomly chosen with a number of cited papers as 4. Our system generated an informative query \textit{Multimodal Research: Image-Text Model Interaction} and retrieved relevant papers where the top recommended paper was also cited in the original paper. While zero-shot generation provides valuable insights into existing literature, plan-based generation produces a more succinct and readily usable literature review. 

\section{Conclusion and Future Work}

In this work, we introduce and describe LitLLM, a system which can generate literature reviews in a few clicks from an abstract using off-the-shelf LLMs. This LLM-powered toolkit relies on the RAG with a re-ranking strategy to generate a literature review with attribution. Our auxiliary tool allows researchers to actively search for related work based on a preliminary research idea, research proposal or even a full abstract. We present a modular pipeline that can be easily adapted to include the next generation of LLMs and other domains, such as news, by changing the source of retrieval information. 

Given the growing impact of different LLM-based writing assistants, we are optimistic that our system may aid researchers in searching relevant papers and improve the quality of automatically generated related work sections of a paper. While our system shows promise as a helpful research assistant, we believe that their usage should be disclosed to the readers, and authors should also observe caution in eliminating any possible hallucinations. 

In the future, we would also like to explore academic search through multiple APIs, such as Google Scholar. This work only considered abstracts of the query paper and the retrieved papers, which creates a bottleneck in effective literature review generation. With the advent of longer context LLMs, we envision our system ingesting the whole paper (potentially leveraging an efficient LLM-based PDF parser) to provide a more relevant background of the related research. We consider our approach as an initial step for building intelligent research assistants which could help academicians through an interactive setting \citep{dwivedi2022editeval}.

\bibliography{acl_latex}

\appendix


\section*{Appendix}

\begin{figure}[ht]
\centering
\includegraphics[width=\linewidth]{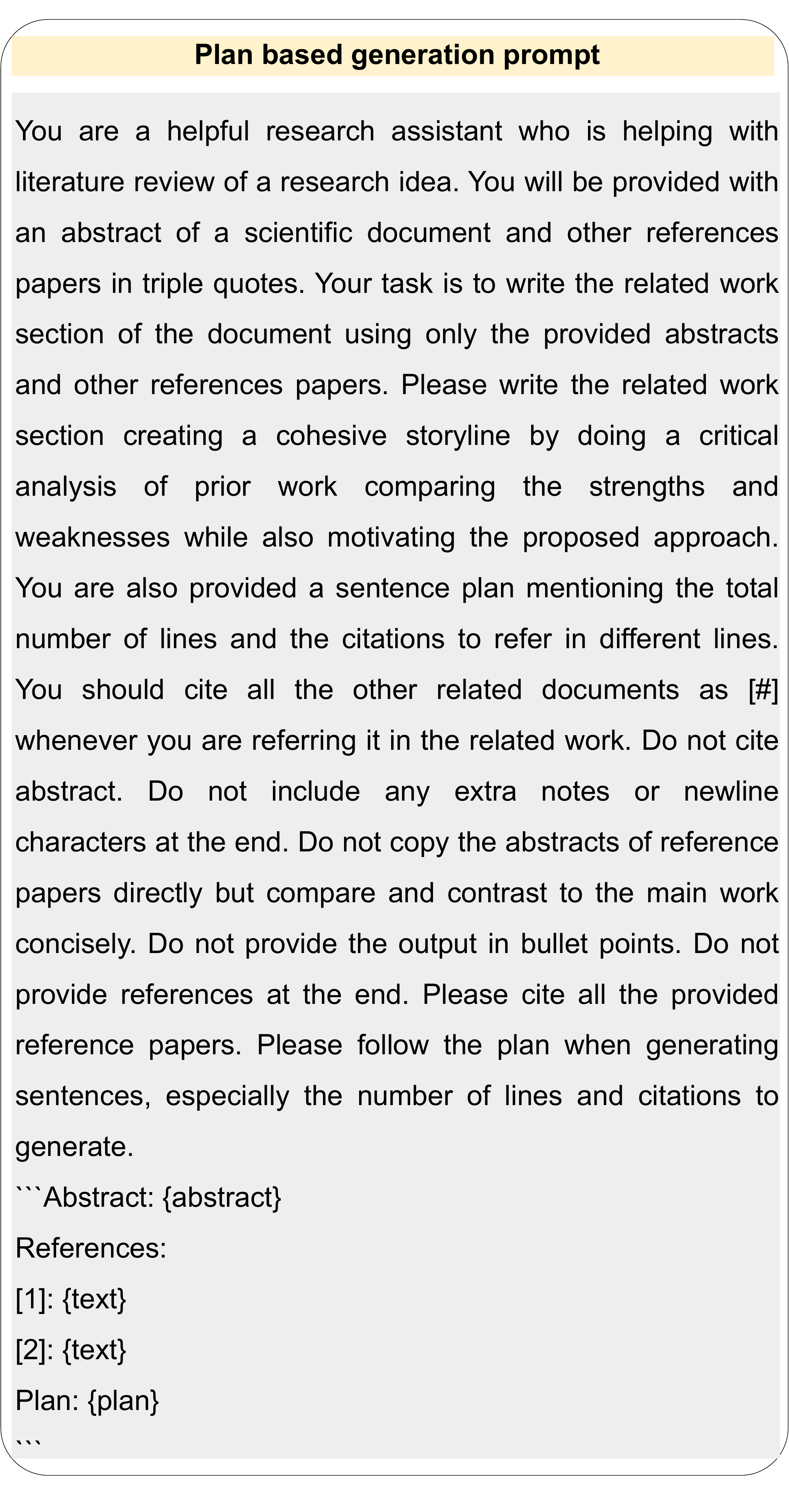}
\caption{Prompt for sentence plan-based generation}
\label{fig:plan-generation-prompt}
\end{figure} 

In the following, we provide snippets of code to retrieve results from the Semantic Scholar API for both recommendation and query-based search:

\begin{lstlisting}[language=python]
# QUERY BASED SEARCH

def query_search_s2(query: str, 
                    num_papers_api: int, 
                    fields: str):
    rsp = requests.get("https://api.semanticscholar.org/graph/v1/paper/search", 
    headers={"X-API-KEY": S2_API_KEY}, 
    params={"query": query, "limit": num_papers_api, "fields": fields})
    rsp.raise_for_status()
    results = rsp.json()
    # Total papers found
    total = results["total"] 
    papers = results["data"]
    return papers
\end{lstlisting}

\begin{lstlisting}[language=python]
# PAPER SEARCH

def get_paper_data(paper_url: str, 
                   fields: str):
    """
    Retrieves data of one paper based on URL
    """
    rsp = requests.get(f"https://api.semanticscholar.org/graph/v1/paper/URL:{paper_url}",
    headers={"X-API-KEY": S2_API_KEY},
    params={"fields": fields})
    results = rsp.json()
    return rewsults
    
\end{lstlisting}

\begin{lstlisting}[language=python]
# RECOMMENDATION API

def get_recommendations_from_s2(
    arxiv_id: str, num_papers_api: int, 
    fields: str):
    """
    Get recommendations from S2 API
    """
    query_id = f"ArXiv:{arxiv_id}"
    rsp = requests.post("https://api.semanticscholar.org/recommendations/v1/papers/",
    json={
        "positivePaperIds": [query_id]},
    params={"fields": fields, 
            "limit": num_papers_api})
    results = rsp.json()
    papers = results["recommendedPapers"]
    return papers
    
\end{lstlisting}




\begin{table*}[ht]
\centering
\resizebox{\textwidth}{!}{ 
\scriptsize{ 
 \begin{tabular}{p{0.9\textwidth}}
  \hline
  \textbf{Abstract for paper: \textit{Do DALL-E and Flamingo Understand Each Other? ~\cite{li2023dall}}}\\\hline 		 
The field of multimodal research focusing on the comprehension and creation of both images and text has witnessed significant strides. This progress is exemplified by the emergence of sophisticated models dedicated to image captioning at scale, such as the notable Flamingo model and text-to-image generative models, with DALL-E serving as a prominent example. An interesting question worth exploring in this domain is whether Flamingo and DALL-E understand each other. To study this question, we propose a reconstruction task where Flamingo generates a description for a given image and DALL-E uses this description as input to synthesize a new image. We argue that these models understand each other if the generated image is similar to the given image. Specifically, we study the relationship between the quality of the image reconstruction and that of the text generation. We find that an optimal description of an image is one that gives rise to a generated image similar to the original one. The finding motivates us to propose a unified framework to finetune the text-to-image and image-to-text models. Concretely, the reconstruction part forms a regularization loss to guide the tuning of the models. Extensive experiments on multiple datasets with different image captioning and image generation models validate our findings and demonstrate the effectiveness of our proposed unified framework. As DALL-E and Flamingo are not publicly available, we use Stable Diffusion and BLIP in the remaining work. Project website: \url{https://dalleflamingo.github.io.}
\\ \hline 
\textbf{LLM summarized query:}
Multimodal Research: Image-Text Model Interaction \\
\textbf{Retrieved papers:} (User Input: 4) \\
$[1]$ CoCa: Contrastive Captioners are Image-Text Foundation Models (2022) Cited by 702 ~\citep{Yu2022CoCaCC} \\
$[2]$ MAMO: Fine-Grained Vision-Language Representations Learning with Masked Multimodal Modeling (2022) Cited by 0 ~\citep{Zhao2022MAMOFV} \\
$[3]$ Dynamic Modality Interaction Modeling for Image-Text Retrieval (2021) Cited by 88 ~\citep{Qu2021DynamicMI} \\
$[4]$ WIT: Wikipedia-based Image Text Dataset for Multimodal Multilingual Machine Learning (2021) Cited by 185 ~\citep{Srinivasan2021WITWI} 
\\ \hline 
\textbf{Zero shot Generation (Section \ref{subsec:zero})} \\ \hline
The field of multimodal research has seen significant advancements in recent years, with the development of models such as Flamingo and DALL-E that focus on image captioning and text-to-image generation respectively. However, the question of whether these models can understand each other and work in harmony is a topic of interest. In this context, the work of \textcolor{blue}{[1]} presents the Contrastive Captioner (CoCa), a model that combines contrastive loss and captioning loss to pretrain an image-text encoder-decoder foundation model. This model, while efficient, does not fully address the interaction between image and text modalities at a fine-grained level.

The work of \textcolor{purple}{[2]} addresses this gap by proposing a jointly masked multimodal modeling method that focuses on fine-grained multimodal representations. This method performs joint masking on image-text input and integrates both implicit and explicit targets for the masked signals to recover. However, this approach, while effective, does not fully address the challenges of intra-modal reasoning and cross-modal alignment that are inherent in image-text retrieval.

The work of \textcolor{orange}{[3]} presents a novel modality interaction modeling network based on the routing mechanism, which is the first unified and dynamic multimodal interaction framework towards image-text retrieval. This model can dynamically learn different activated paths for different data, providing a more flexible approach to modality interaction. However, the design of interaction patterns in this model still relies heavily on expert experience and empirical feedback, which may limit its applicability in different contexts.
The work of \textcolor{red}{[4]} introduces the Wikipedia-based Image Text (WIT) Dataset, a large-scale dataset for multimodal, multilingual learning. This dataset, while providing a rich resource for multimodal learning, does not directly address the question of how different models can understand each other and work together.

In light of these previous works, our study proposes a reconstruction task where Flamingo generates a description for a given image and DALL-E uses this description as input to synthesize a new image. We argue that these models understand each other if the generated image is similar to the given image. This approach allows us to study the relationship between the quality of the image reconstruction and that of the text generation, and to propose a unified framework to finetune the text-to-image and image-to-text models. Our extensive experiments validate our findings and demonstrate the effectiveness of our proposed unified framework.
\\\hline
\textbf{Plan based generation (Section \ref{subsec:plan})}: Generate the output using 5 sentences. Cite [1] on line 2. Cite [2], [3] on line 3. Cite [4] on line 5. \\ \hline
The field of multimodal research has seen significant advancements in the comprehension and creation of both images and text, with models like Flamingo and DALL-E leading the way. In a similar vein, the Contrastive Captioner (CoCa) model presented in \textcolor{blue}{[1]} pretrains an image-text encoder-decoder foundation model with contrastive loss and captioning loss, achieving state-of-the-art performance on a broad range of downstream tasks. Other works have also focused on improving the fine-grained image-text interaction, with \textcolor{purple}{[2]} proposing a jointly masked multimodal modeling method and \textcolor{orange}{[3]} developing a dynamic multimodal interaction framework for image-text retrieval. These models, while effective, often rely on expert experience and empirical feedback, which may limit their flexibility. The introduction of the Wikipedia-based Image Text (WIT) Dataset in \textcolor{red}{[4]} provides a large, multilingual dataset for pretraining multimodal models, offering a more diverse set of concepts and real-world entities for model training. This diversity and scale could potentially enhance the performance of models like Flamingo and DALL-E, further advancing the field of multimodal research.
\\\hline
  \end{tabular}
  }}
\caption{We show an example generated related work for a randomly chosen recent paper~\cite{li2023dall} with LLM summarized query and retrieved papers. We show the generated related work from our system using both zero-shot and plan-based generation, producing a more succinct and readily usable literature review. Note: The number of citations is retrieved by Semantic Scholar at the date of submission of this work.}
\label{table:qualitative}
\end{table*}

\end{document}